\documentclass[10pt,twocolumn,letterpaper]{article}

\usepackage{wacv}
\usepackage{times}
\usepackage{epsfig}
\usepackage{graphicx}
\usepackage{amsmath}
\usepackage{amssymb}
\usepackage[caption=false]{subfig}
\newcommand{\Mf}[0]{\mathcal{M}}
\newcommand{\Md}[0]{\mathcal{M}_\text{ip}}
\newcommand{\Mc}[0]{\mathcal{M}_\text{cam}}
\newcommand{\I}[0]{\mathbf{I}}
\renewcommand{\P}[0]{\mathbf{P}}  
\newcommand{\M}[0]{\mathbf{M}}
\newcommand{\Dtip}[0]{\mathcal{D}_\text{T}^\text{ip}}
\newcommand{\Dtcam}[0]{\mathcal{D}_\text{T}^\text{cam}}
\newcommand{\Dt}[0]{\mathcal{D}_\text{T}}
\newcommand{\Dv}[0]{\mathcal{D}_\text{V}}
\newcommand{\De}[0]{\mathcal{D}_\text{E}}

\wacvfinalcopy 

\def\wacvPaperID{248} 

\ifwacvfinal\fi
\begin{document}

\title{Reliability Map Estimation For CNN-Based Camera Model Attribution}

\author{David G\"uera\\
Purdue University\\
West Lafayette, Indiana\\
\and
Sri Kalyan Yarlagadda\\
Purdue University\\
West Lafayette, Indiana\\
\and
Paolo Bestagini\\
Politecnico di Milano\\
Milan, Italy\\
\and
Fengqing Zhu\\
Purdue University\\
West Lafayette, Indiana\\
\and
Stefano Tubaro\\
Politecnico di Milano\\
Milan, Italy\\
\and
Edward J. Delp\\
Purdue University\\
West Lafayette, Indiana\\
}

\maketitle
\ifwacvfinal\thispagestyle{empty}\fi

\begin{abstract}
Among the image forensic issues investigated in the last few years, great attention has been devoted to blind camera model attribution. This refers to the problem of detecting which camera model has been used to acquire an image by only exploiting pixel information. Solving this problem has great impact on image integrity assessment as well as on authenticity verification. Recent advancements that use convolutional neural networks (CNNs) in the media forensic field have enabled camera model attribution methods to work well even on small image patches. These improvements are also  important for determining forgery localization. Some patches of an image may not contain enough information related to the camera model (e.g., saturated patches).
In this paper, we propose a CNN-based solution to estimate the camera model attribution reliability of a given image patch.
We show that we can estimate a reliability-map indicating which portions of the image contain reliable camera traces.
Testing using a well known dataset confirms that by using this information, it is possible to increase small patch camera model attribution accuracy by more than $8\%$ on a single patch.
\end{abstract}


\section{Introduction}\label{sec:intro}

Due to the widespread availability of inexpensive image capturing devices (e.g., cameras and smartphones) and user-friendly editing software (e.g., GIMP and Adobe PhotoShop), image manipulation is very easy. For this reason, the multimedia forensic community has developed techniques for image authenticity detection and integrity assessment \cite{Rocha2011a, Piva2013, Stamm2013}. 

Among the problems considered in the forensic literature, one important problem is camera model attribution, which consists in estimating the camera model used to acquire an image \cite{Kirchner2015}. This proves useful when a forensic analyst needs to link an image under investigation to a user \cite{Filler2008}, or to detect possible image manipulations \cite{Swaminathan2008, Cozzolino2014} (e.g., splicing of pictures from different cameras).

Linking an image to a camera can in principle be trivially done exploiting image header information (e.g., EXIF data). It is also true that image headers are not reliable (e.g., anyone can tamper with them) or not always available (e.g., decoded images and screen captures). Therefore, the need for a series of blind methodologies has led to the development of pixel-based only information extraction methods.

\begin{figure}[t]
	\centering
	\includegraphics[width=\columnwidth]{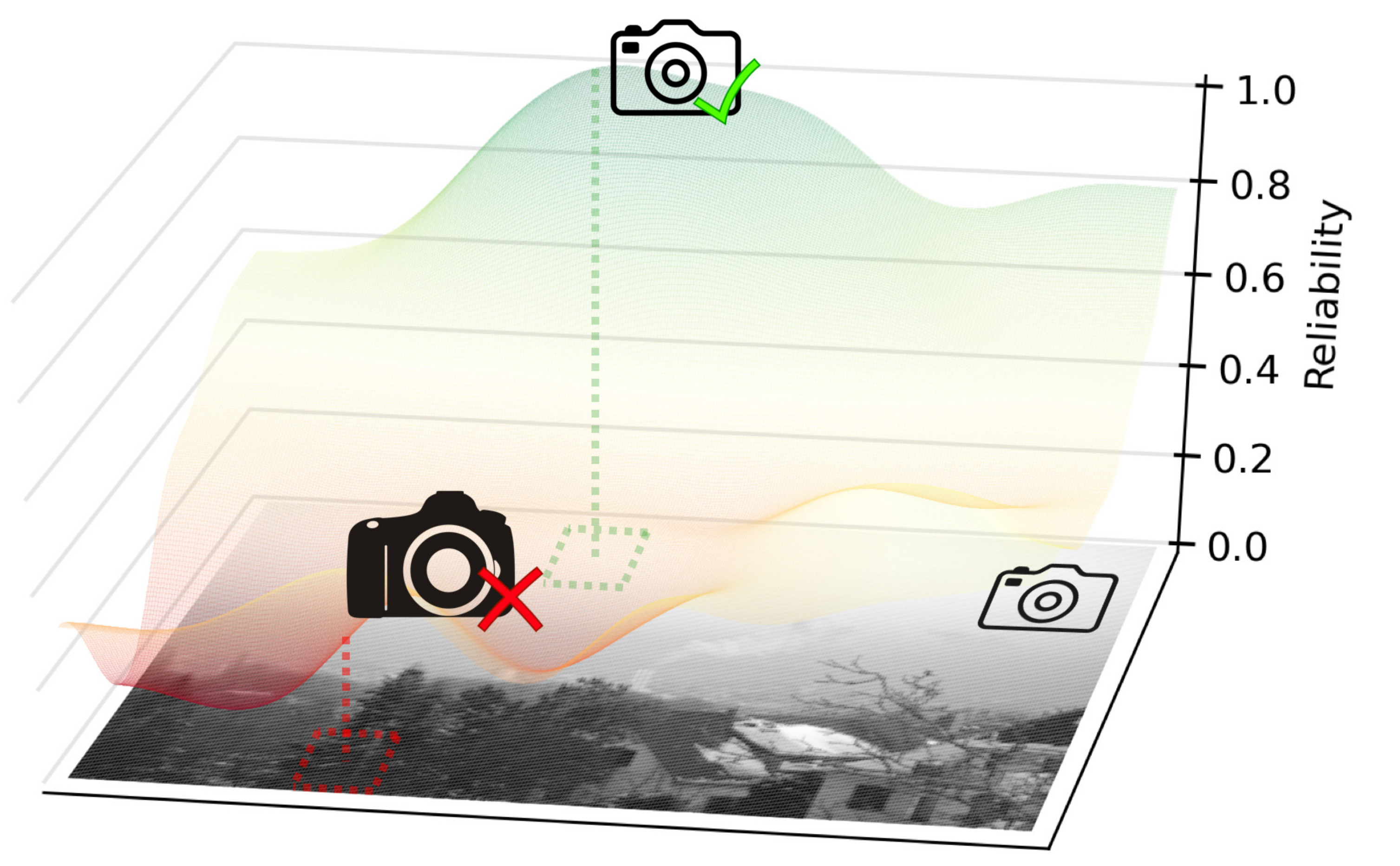}
	\caption{Reliability map representation for an example image taken with a given camera. In this case, patches belonging to the sky (green box) are more likely to provide accurate camera model attribution than patches containing textures (red box).}
	\label{fig:reliability_intro}
\end{figure}

These methods leverage the fact that image acquisition pipeline is slightly different for each camera model and manufacturer (e.g., different sensors and color equalization techniques). Therefore, each image contains characteristic ``fingerprints'' that enable one to understand which pipeline has been used and hence the camera model. 
Among these techniques,  exploiting photo sensor non uniformity (PRNU) is particularly robust and enables camera instance identification \cite{Lukas2006, Goljan2008}. 
Other methods exploit traces left by color filter array (CFA) interpolation \cite{Bayram2005, Cao2010, Zhao2016}, camera lenses \cite{Choi2006}, histogram equalization \cite{Chen2007a} or noise \cite{Thai2014}. Alternatively, a series of methods extracting statistical features from the pixel-domain and exploiting supervised machine-learning classifier have also been proposed \cite{Chen2015a, Tuama2016, Marra2015}.

Due to the advancements brought by deep learning techniques in the last few years, the forensic community is also  exploring convolutional neural networks (CNNs) for camera model identification \cite{Tuama2016a}. Interestingly, the approach in \cite{Bondi2017} has shown the possibility of accurately estimating the camera model used to acquire an image by analyzing a small portion of the image (i.e., a $64\times64$ color image patch). This has lead to the development of forgery localization techniques \cite{Bondi2017b}.

In this paper we propose a CNN-based method for estimating patch reliability for camera model attribution. As explained in \cite{Bondi2017a}, not all image patches contain enough discriminative information to estimate the camera model (e.g., saturated areas and too dark regions). Leveraging the network proposed in \cite{Bondi2017}, we show how it is possible to determine whether an image patch contains reliable camera model traces for camera model attribution. Using this technique, we build a reliability map, which indicates the likelihood of each image region to be possibly used for camera model attribution, as shown in Figure~\ref{fig:reliability_intro}. This map can be used to select only reliable patches for camera model attribution. Additionally, it can also be used to drive tampering localization methods \cite{Bondi2017b} by providing valuable information on which patches should be considered to be unreliable.

The proposed method leverages CNN feature learning capabilities and transfer learning training strategies. Specifically, we make use of a CNN composed by the architecture proposed in \cite{Bondi2017} as feature extractor, followed by a series of fully connected layers for patch reliability estimation. Transfer learning enables to preserve part of the CNN weights of \cite{Bondi2017}, and train the whole architecture end-to-end with a reduced number of image patches. Our strategy is validated on the  Dresden Image Database \cite{Gloe2010}. We first validate the proposed architecture and training strategy. Then, we compare the proposed solution against a set of baseline methodologies based on classic supervised machine-learning techniques. Finally, we show how it is possible to increase camera model attribution accuracy by more than $8\%$ with respect to \cite{Bondi2017} using the proposed method.

\section{Problem Statement and Related Work}\label{sec:problem}
In this section we introduce the problem formulation with the notation used throughout the paper. We then provide the reader a brief overview about CNNs and their use in multimedia forensics.

\subsection{Problem Formulation}
Let us consider a color image $\I$ acquired with camera model $l$ belonging to a set of known camera models $\mathcal{L}$. In this paper, we consider the patch-based closed-set camera model attribution problem as presented in \cite{Bondi2017}. Given an image $\I$, this means
\begin{itemize}
	\item Select a subset of $K$ color patches $\P_k, \; k \in [1, K]$.
	\item Obtain an estimate $\hat{l}_k = \mathcal{C}(\P_k)$ of the camera model associated with each patch through a camera attribution function $\mathcal{C}$.
	\item Optionally obtain final camera model estimate $\hat{l}$ through majority voting over $\hat{l}_k, \; k \in [1, K]$.
\end{itemize}

Our goal is to detect whether a patch $\P_k$ is a good candidate for camera model attribution estimation. To this purpose, we propose a CNN architecture that learns a function $\mathcal{G}$ expressing the likelihood of a patch $\P_k$ to provide correct camera model identification, i.e., $g_k = \mathcal{G}(\P_k)$. High values of $g_k$ indicate high probability of patch $\P_k$ to provide correct camera information. Conversely, low $g_k$ values are attributed to patches $\P_k$ that cannot be correctly classified. Pixel-wise likelihood is then represented by means of a reliability map $\M$, showing which portion of an image is a good candidate to estimate image camera model, as shown in Figure~\ref{fig:reliability_intro}.

\subsection{Convolutional Neural Networks in Multimedia Forensics}
In this section, we present a brief overview of the foundations of convolutional neural networks (CNNs) that are needed to follow the paper. 
For a thorough review on CNNs, we refer the readers of this paper to Chapter 9 of \cite{Goodfellow2016}. 

Deep learning and in particular CNNs have shown very good performance in several computer vision applications such as visual object recognition, object detection and many other domains such as drug discovery and genomics \cite{LeCun2015}. Inspired by how the human vision works, the layers of a convolutional network have neurons arranged in three dimensions, so each layer has a width height, and depth. The neurons in a convolutional layer are only connected to a small, local region of the preceding layer, so we avoid wasting resources as it is common in fully-connected neurons. The nodes of the network are organized in multiple stacked layers, each performing a simple operation on the input. The set of operations in a CNN typically comprises convolution, intensity normalization, non-linear activation and thresholding, and local pooling. By minimizing a cost function at the output of the last layer, the weights of the network are tuned so that they are able to capture patterns in the input data and extract distinctive features. 
CNNs enable learning data-driven, highly representative, layered hierarchical image features from sufficient training data

\begin{figure*}[t]
	\centering
	\includegraphics[width=\textwidth]{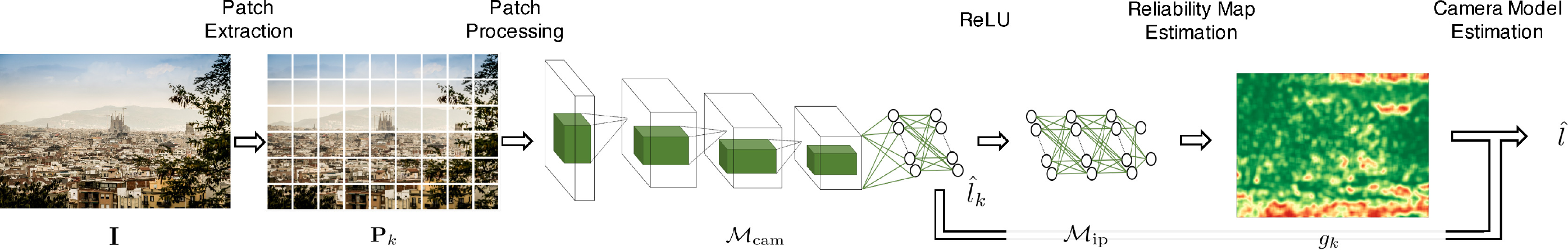} \vspace{.1em}
	\caption{Block diagram of the proposed approach. Image $\I$ is split into patches. Each patch $\P_k, \; k \in [0, K]$ is processed by the proposed CNN (composed by $\Mc$ and $\Md$) to obtain a reliability score $g_k$ and a camera model estimate $\hat{l}_k$. The reliability map is determined from all $g_k, \; k \in [0, K]$ values, and the overall picture camera model estimate $\hat{l}$ can be computed.}
	\label{fig:block_diagram}
\end{figure*}

To better understand the role of each layer, we describe the most common building blocks in a CNN:
\begin{itemize}
	\item \emph{Convolution}: each convolution layer is a filterbank, whose filters impulse response $h$ are learned through training. Given an input signal $x$, the output of each filter is $y=x*h$, i.e., the valid part of the linear convolution. Convolution is typically done on 3D representations consisting of the spatial coordinates $(x, y)$ and the number of feature maps $p$ (e.g., $p = 3$ for an RGB input).
	\item \emph{Max pooling}: returns the maximum value of the input $x$ evaluated across small windows (typically of 3x3 pixels).
	\item \emph{ReLU}: Rectified Linear Units use the rectification function $y=\max(0, x)$ to the input $x$, thus clipping negative values to zero.
	\item \emph{Inner Product}: indicates that the input of each neuron of the next layer is a linear combination of all the outputs of the previous layer. Combination weights are estimated during training. 
	\item \emph{SoftMax}: maps the input into compositional data (i.e., each value is in the range $[0, 1]$, and they all sum up to one). This is particularly useful at the end of the network in order to interpret its outputs as probability values.
\end{itemize}

There has been a growing interest in using convolutional neural networks in the fields of image forensics and steganalysis \cite{Xu2016, Chen2017}. These papers mainly focus on architectural design of CNNs where a single CNN model is trained and then tested in experiments. Data-driven models have recently proved valuable for other multimedia forensic applications as well \cite{Chen2015, Bayar2016}. Moreover, initial exploratory solutions targeting camera model identification \cite{Tuama2016b, Bondi2017, Bondi2017b} show that it is possible to use CNNs to learn discriminant features directly from the observed known images, rather than having to use hand-crafted features. As a matter of fact, the use of CNNs also makes it possible to capture characteristic traces left by non-linear and hard to model operations present in the image acquisition pipeline of capturing devices.

In this paper, we employ CNNs as base learners and test several different training strategies and network topologies. In our study, at first, a recently proposed CNN architecture is adopted as a feature extractor, trained on a random subsample of the training dataset. An intermediate feature representation is then extracted from the original data and pooled together to form new features ready for the second level of classification. Results have indicated that learning from intermediate representation in CNNs instead of output probabilities, and then jointly retraining the final architecture, leads to performance improvement.

\section{Patch Reliability Estimation Method}\label{sec:algorithm}
In this section we provide details of our method for patch reliability estimation and camera attribution. The proposed pipeline is composed by the following steps (see Figure~\ref{fig:block_diagram}):
\begin{enumerate}
	\item The image under analysis is split into patches
	\item A CNN is used to estimate patch reliability likelihood
	\item From the same CNN we estimate a camera model for each patch
	\item A reliability mask is constructed and camera attribution of the whole image is performed
\end{enumerate}
Below is a detailed explanation of each step.

\begin{figure*}[t]
	\centering
	\includegraphics[width=.9\textwidth]{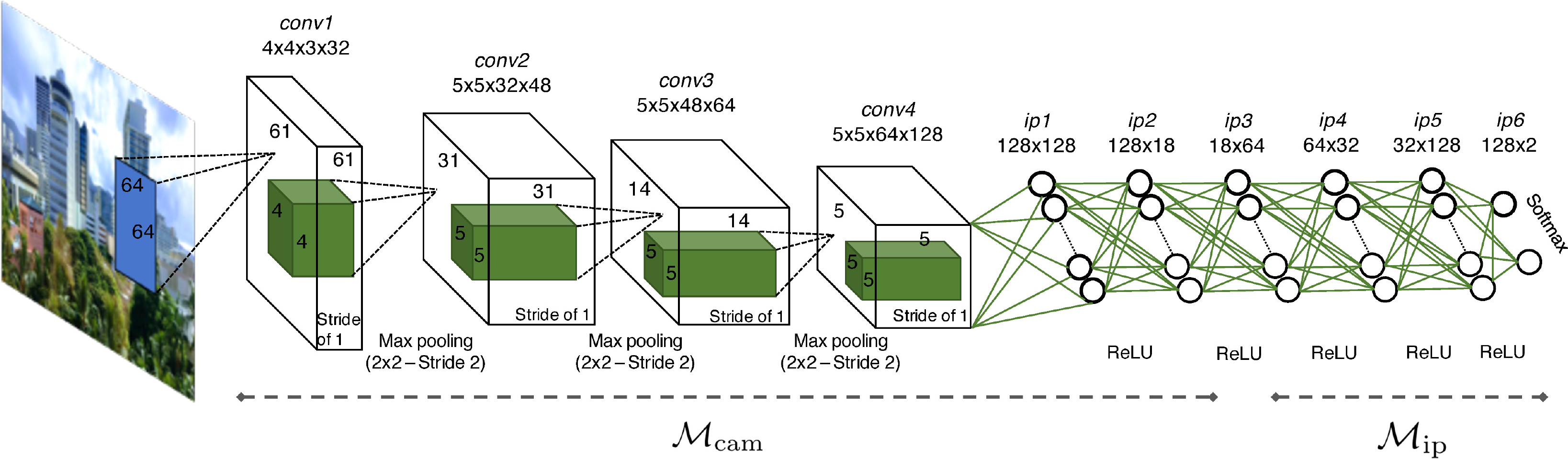}
	\caption{Representation of the proposed CNN architecture $\Mf$ working on image patches. The first part ($\Mc$) computes a $|\mathcal{L}|$-element feature vector used for camera attribution. The second part ($\Md$) is used to derive the camera model attribution reliability.}
	\label{fig:cnn}
\end{figure*}

\subsection{Patch Extraction}
The proposed method works by analyzing image patches. The first step is to split the color image $\I$ into a set of $K$ patches $\P_k, \; k \in [0, K]$. Each patch has $64 \times 64$ pixel resolution. The patch extraction stride can range from $1$ to $64$ per dimension, depending on the amount of desired overlap. This can be chosen to balance the trade-off between mask resolution reliability and computational burden.

\subsection{Patch Camera Reliability}
Each patch $\P_k$ is input into the CNN $\Mf$ shown in Figure~\ref{fig:cnn}, which can be logically split into two parts ($\Mc$ and $\Md$) connected through a ReLU activation layer. Our proposed CNN learns a patch reliability function $\mathcal{G}$ and returns the patch reliability $g_k = \mathcal{G}(\P_k)$. 

The first part (i.e., $\Mc$) is the CNN presented in \cite{Bondi2017} without last layer's activation. The rationale behind this choice is that this network is already known to be able to extract characteristic camera information. Therefore, we can mainly think of this portion of the proposed CNN as the feature extractor, turning a patch $\P_k$ into a feature vector in $\mathbb{R}^{|\mathcal{L}|}$, where $|\mathcal{L}|$ is the number of considered camera models. Formally, $\Mc$ is composed by:
\begin{itemize}
\item \textit{conv1}: convolutional layer with 32 filters of size $4\times4\times3$ and stride 1.
\item \textit{conv2}: convolutional layer with 48 filters of size $5\times5\times32$ and stride 1.
\item \textit{conv3}: convolutional layer with 64 filters of size $5\times5\times48$ and stride 1.
\item \textit{conv4}: convolutional layer with 128 filters of size $5\times5\times64$ and stride 1, which outputs a vector with 128 elements.
\item \textit{ip1}: inner product layer with 128 output neurons followed by a ReLU layer to produce a 128 dimensional feature vector.
\item \textit{ip2}: final $128 \times |\mathcal{L}|$ inner product layer.
\end{itemize}
The first three convolutional layers are followed by max-pooling layers with $2\times2$ kernels and $2\times2$ stride. This network contains $360\,462$ trainable parameters.

The second part of our architecture (i.e., $\Md$) is composed by a series of inner product layers followed by ReLU activations. This part of the proposed CNN can be considered as a two-class classifier trying to distinguish between patches that can be correctly classified, and patches that cannot correctly be attributed to their camera model. As shall be clear in Section~\ref{sec:setup}, we tested different possible $\Md$ architecture candidates, to decide upon the following structure (denoted later on as $\Md^4$ due to the 4 layers that characterize it):
\begin{itemize}
	\item \textit{ip3}: inner product layer with 64 output neurons followed by ReLU
	\item \textit{ip4}: inner product layer with 32 output neurons followed by ReLU
	\item \textit{ip5}: inner product layer with 128 output neurons followed by ReLU
	\item \textit{ip6} inner product layer with 2 output neurons followed by softmax normalization
\end{itemize}
The first element of the softmax output vector can be considered the likelihood of a patch to be correctly classified. Therefore, we consider this value as $g_k$, and the transfer function learned by the whole $\Mf$ network as $\mathcal{G}$.

\subsection{Patch Camera Attribution}
In order to detect the camera model from each patch $\P_k$, we exploit the architecture $\Mc$. As explained in \cite{Bondi2017}, this CNN output is a $|\mathcal{L}|$-element vector, whose argmax indicates the used camera model $\hat{l}_k \in \mathcal{L}$. Note that the softmax normalization at the end of $\Mc$ (as proposed in \cite{Bondi2017}) is not needed in this situation, as it only impacts the training strategy and not the argmax we are interested in.

\subsection{Reliability-Map and Camera Attribution}
\begin{figure}[t]
	\centering
	\subfloat{\includegraphics[height=1.1in]{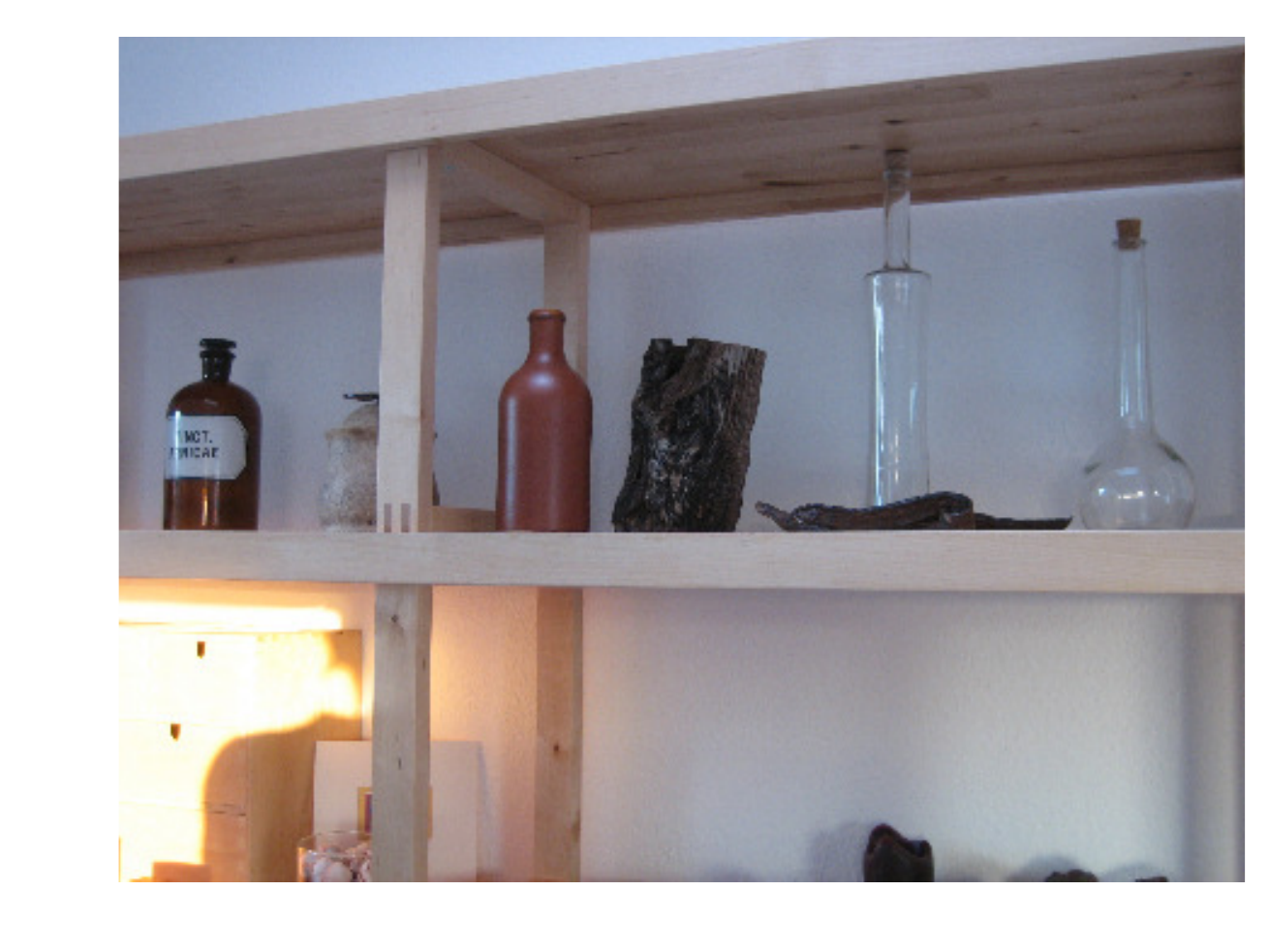}} \quad
	\subfloat{\includegraphics[height=1.1in]{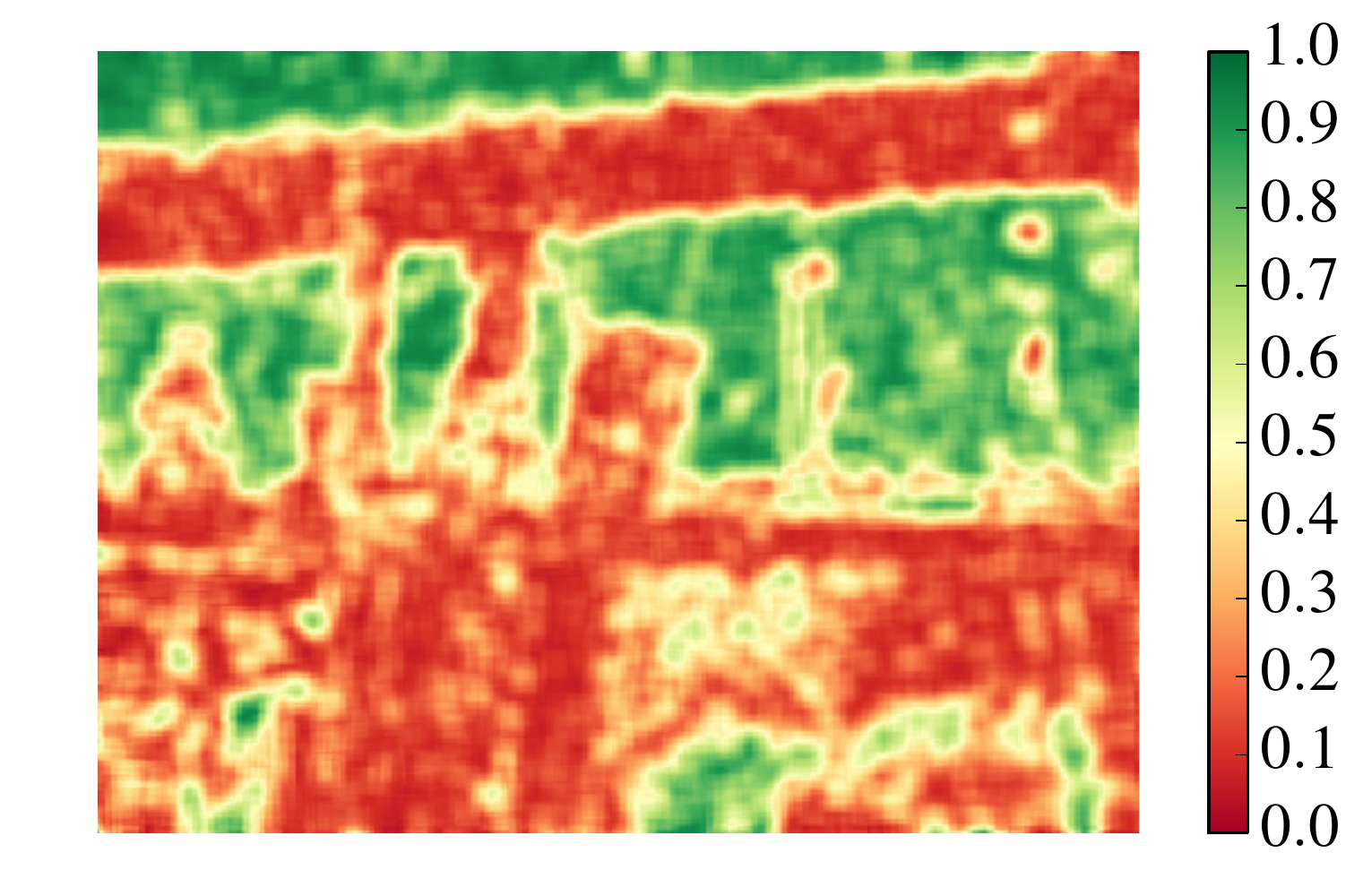}} \\ \vspace{-1em}
	\subfloat{\includegraphics[height=1.1in]{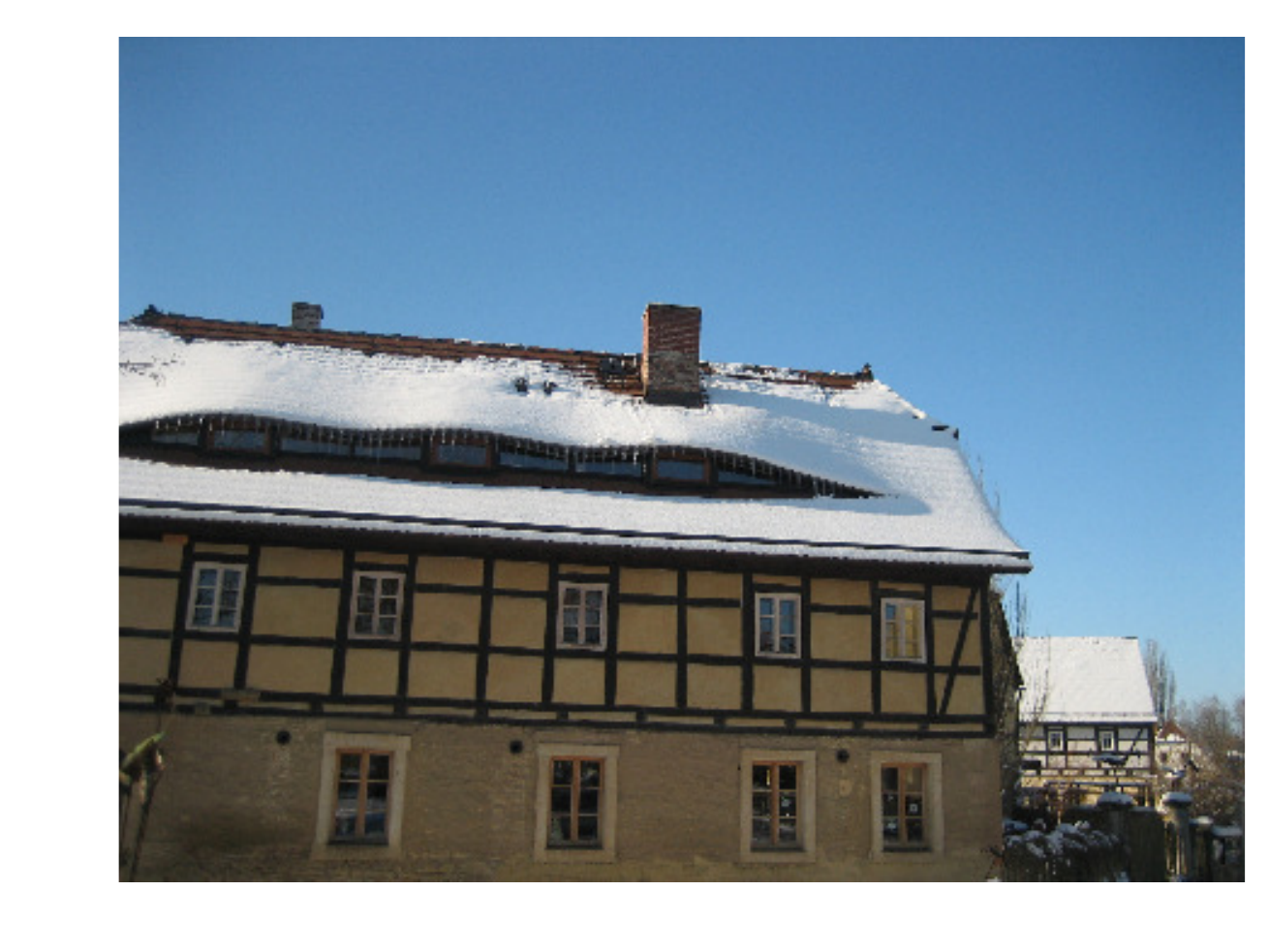}} \quad
	\subfloat{\includegraphics[height=1.1in]{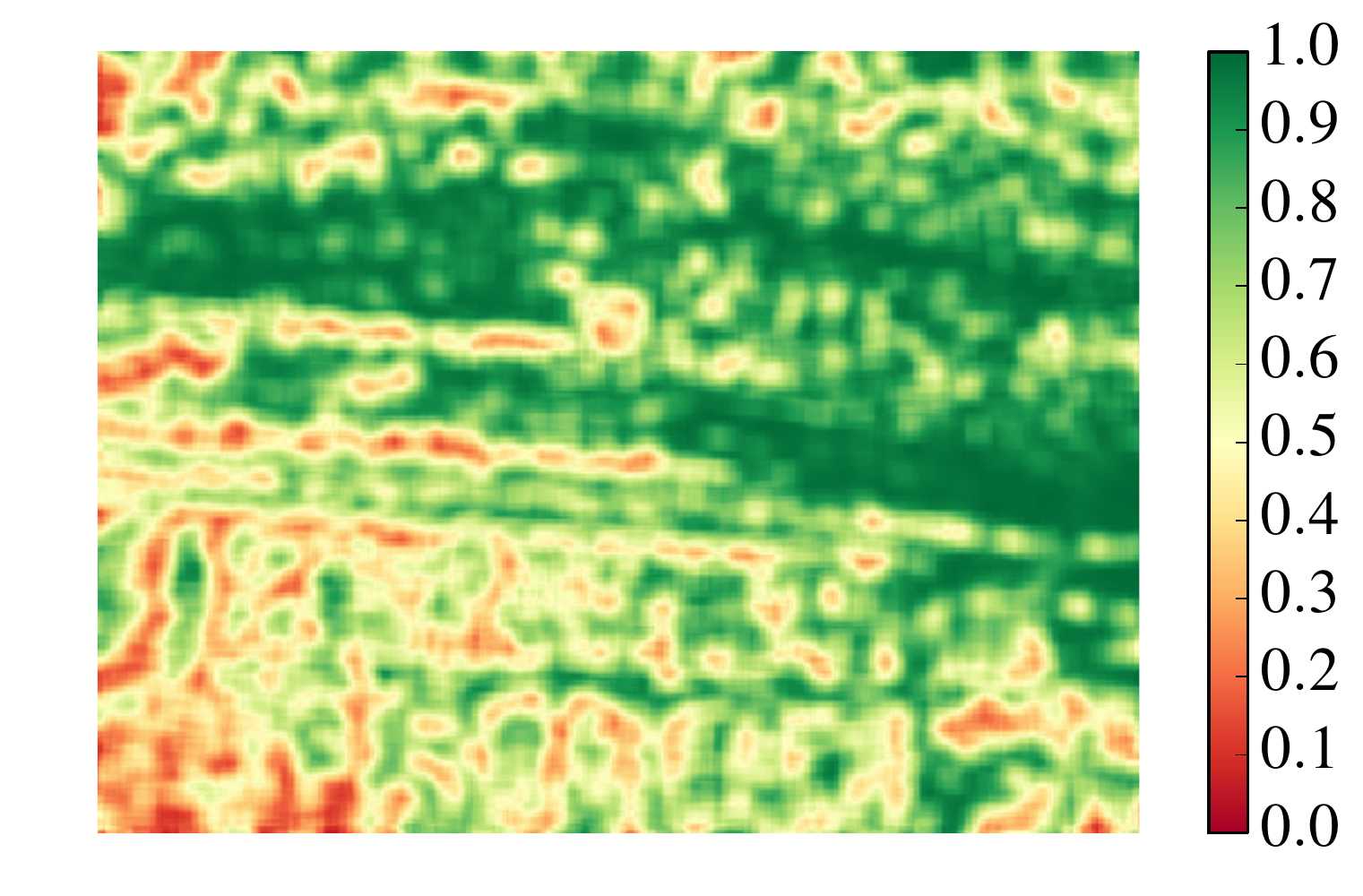}} \\ \vspace{-1em}
	\subfloat[$\I$]{\includegraphics[height=1.1in]{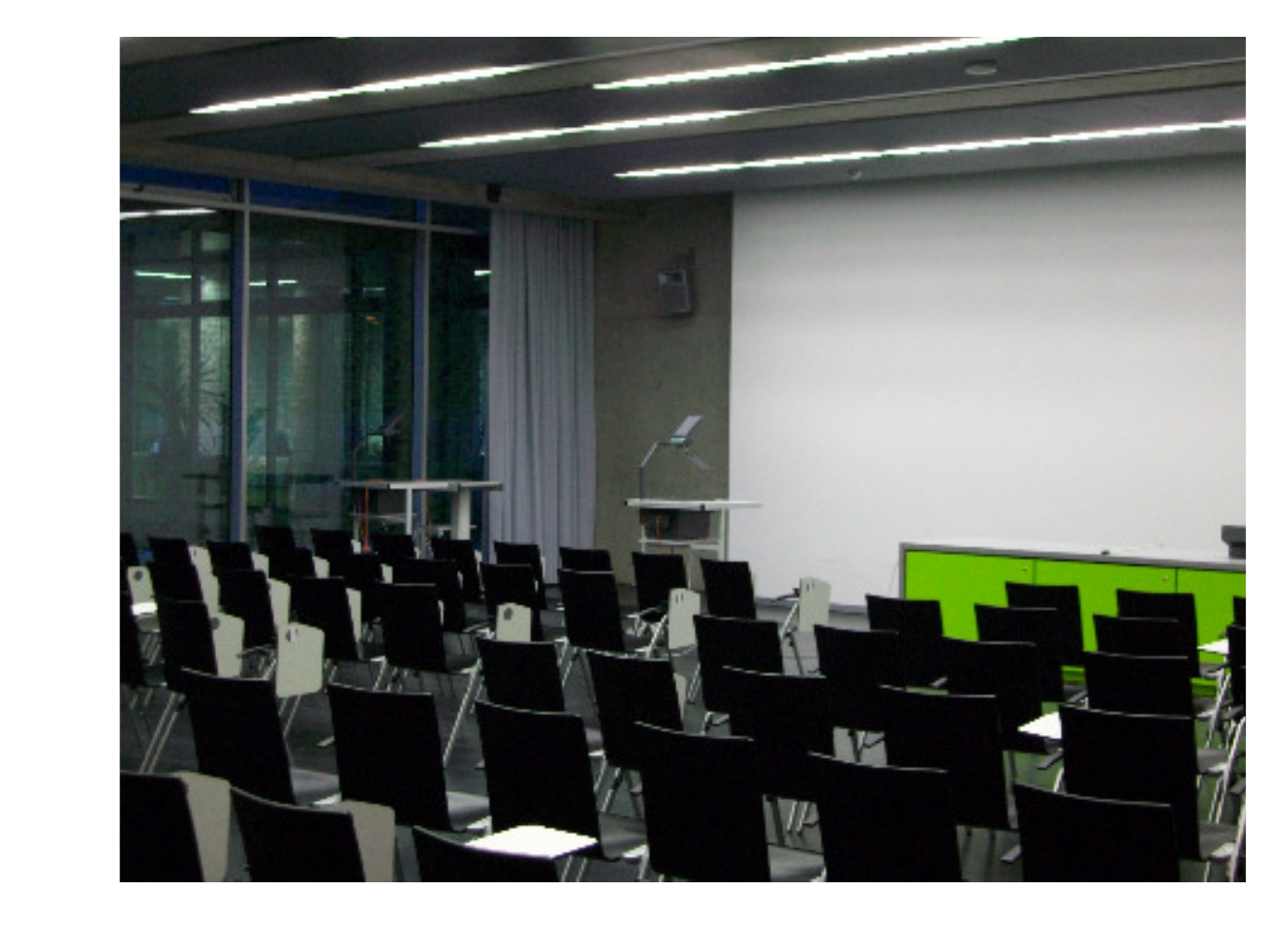}} \quad
	\subfloat[$\M$]{\includegraphics[height=1.1in]{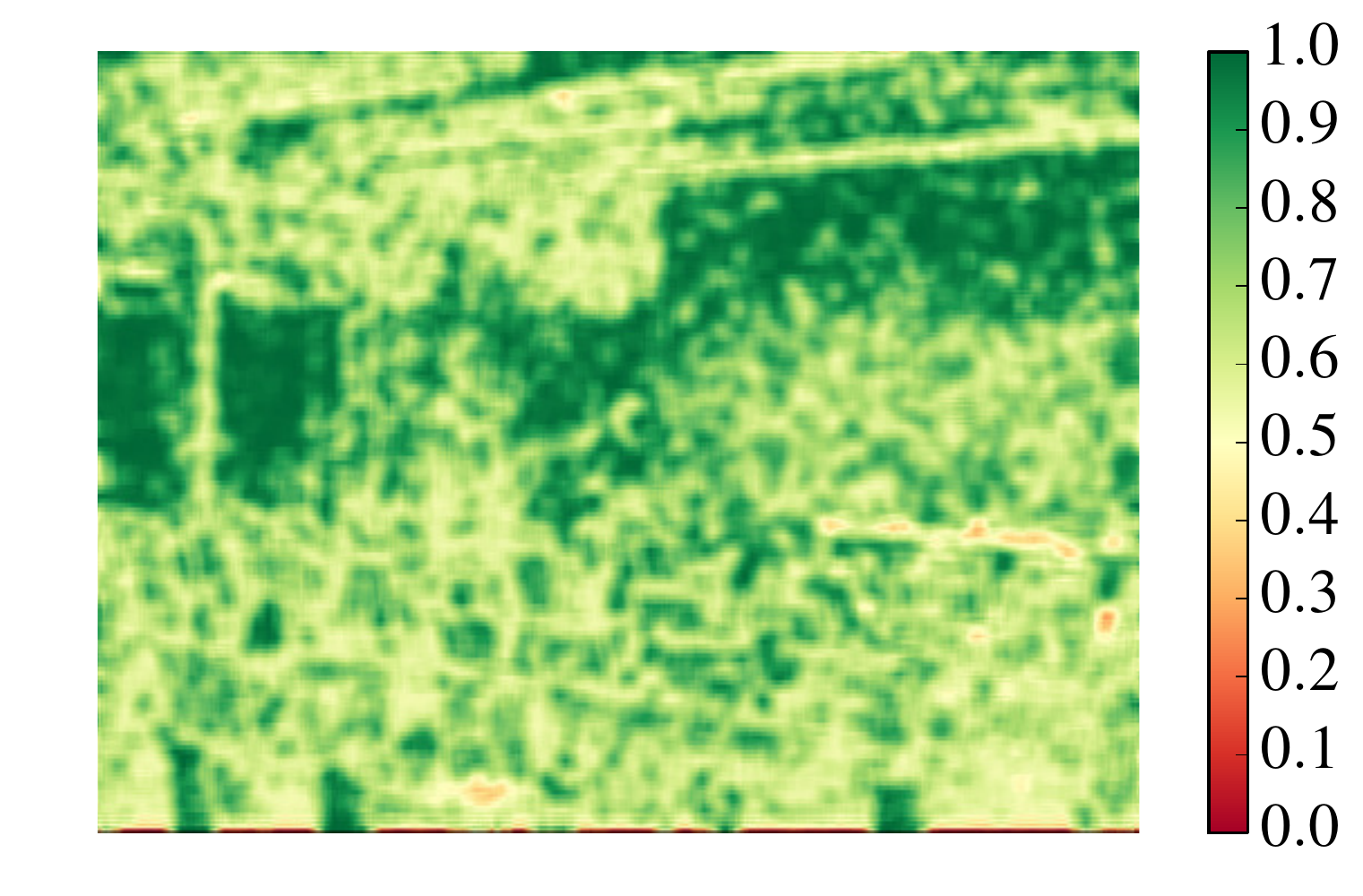}}
	\caption{Examples of images $\I$ (left) and the estimated reliability maps $\M$ (right). Patch reliability is not always strictly linked to the image semantic content. Green areas represent high $g_k$ values, thus reliable regions.}
	\label{fig:masks}
\end{figure}

In order to compute the reliability map $\M$, we aggregate all $g_k$ values estimated from patches $\P_k, \; k \in [1, K]$. This is done by generating a bidimensional matrix $\M$ with the same size of image $\I$, and fill in the positions covered by the patch $\P_k$ with the corresponding $g_k$ values. In case of overlapping patches, $g_k$ values are averaged. This map provides pixel-wise information about image regions from which reliable patches can be extracted. A few examples of estimated maps $\M$ are reported in Figure~\ref{fig:masks}.

Finally, to attribute image $\I$ to a camera model, it is possible to select $l_k$ only for the most reliable patch (i.e., highest $g_k$), or perform majority voting on estimated $l_k$ values belonging only to reliable regions, i.e., $\{ l_k | g_k > \gamma \}$, where $\gamma \in [0,1]$ is the reliability threshold (set to $0.5$ in this paper).

\section{Experimental Results}\label{sec:setup}
In this section we report the details about our experiments. First, we describe the dataset. Then, we provide an insight on the used training strategies.

\subsection{Dataset}
In this paper we evaluate our solution adapting the dataset splitting strategy proposed in \cite{Kirchner2015, Bondi2017, Bondi2017a} to our problem. This strategy is tailored to the Dresden Image Dataset \cite{Gloe2010}, which consists of 73 devices belonging to 25 different camera models. A variable number of shots are taken with each device. Different motives are shot from each position. We refer to a scene as combination of geographical position with a specific motive. With this definition in mind, the dataset consists of a total of 83 scenes. Since we are trying to classify image patches at the level of camera model rather than instance level, we only consider camera models with more than one instance available. This leads us to a total of 18 camera models (as Nikon D70 and D70s basically differ only in their on-device screen) and nearly $15\,000$ shots. 

In order to evaluate our method we divide the dataset consisting of 18 camera models into 3 sets, namely training $\Dt$, validation $\Dv$ and evaluation $\De$. $\Dt$ is again split into two equal sets $\Dtcam$ and $\Dtip$. $\Dtcam$ is used for training the parameters of $\Mc$, whereas $\Dtip$ is used for training $\Md$.  $\Dv$ is used to decide how many epochs to use during training to avoid overfitting. Finally, $\De$ is used for evaluating the trained network on a disjoint set of images in a fair way.

While training a CNN, it is very important to avoid overfitting the data. In our dataset we have images of different scenes taken by different cameras, and our goal is to learn information about camera model from an image. As $\Dv$ is used to avoid overfitting, it is important that $\Dt$ and $\Dv$ are sufficiently different from each other. It is also important that we test on data that has variation with respect to the training data. In order to achieve these goals, we do the following:
\begin{itemize}
	\item Images for $\De$ are selected from a single instance per camera and a set of 11 scenes.
	\item Images for $\Dt$ are selected from the additional camera instances and 63 different scenes.
	\item Images in $\Dv$ are selected from the same camera instances used for $\Dt$, but from the remaining 10 scenes. This makes sets $\Dv$ and $\Dt$ disjoint with respect to scenes, leading to robust training.
\end{itemize}

For training, validation and testing $K=300$ non overlapping color patches of size $64\times64$ are extracted from each image. the resulting dataset $\Dtcam$ contains more than $500\,000$ patches split into $18$ classes. $\Dtip$ is reduced to $90\,000$ patches to balance reliable and non-reliable image patches according to $\Mc$ classification results. Finally, $\Dv$ and $\De$ are composed by more than $700\,000$ and $800\,000$ patches, respectively.

\subsection{Training Strategies}
Given that the proposed approach builds upon a pre-trained network (i.e., $\Mc$), we propose a two-tiered transfer learning-based approach denoted as \textit{Transfer}. For the sake of comparison, we also test two additional strategies, namely \textit{Scratch} and \textit{Pre-Trained}. In the following we report details about each strategy.

\textbf{Scratch.}
This training strategy is the most simple one. It consists in training the whole two-class architecture $\Mf$ using only $\Dtip$ for training and $\Dv$ for validation. This can be considered as a baseline training strategy. We use Adam optimizer with default parameters as suggested in \cite{Kingma2014} and batch size $128$. Loss is set to binary-crossentropy.

\textbf{Pre-Trained.}
This strategy takes advantage of the possibility of using a pre-trained $\Mc$. In this case, we train $\Mc$ for camera model attribution using softmax normalization on its output. Training is carried out on $\Dtcam$ and validation on $\Dv$. Once $\Mc$ has been trained, we freeze its weights, and train the rest of the architecture $\Md$ as a two-class classifier (i.e., reliable vs. non-reliable patches) using $\Dtip$ and $\Dv$. Optimization during both training steps is carried out using Adam optimizer with default values and batches of $128$ patches. We select categorical-crossentropy as our loss function.

\textbf{Transfer.}
This two-tiered training strategy is meant to fully exploit the transfer learning capability of the proposed architecture. The first step consists in training $\Mc$ for camera model attribution using softmax normalization on its output and $\Dtcam$ and $\Dv$ as datasets. This training step is optimized using Adam with default parameters, $128$ patches per batch, and categorical-crossentropy as loss function.

The second step of $\Mf$ training consists in freezing all the convolutional layers of $\Mc$, and continue training all the inner product layers of both $\Mc$ and $\Md$ using the datasets $\Dtip$ and $\Dv$. This enables to jointly learn the weights of the classifier $\Md$, and tailor feature extraction procedure in the last layers of $\Mc$ (i.e., \textit{ip1} and \textit{ip2}) to the classification task. For this step we use binary-crossentropy as loss, and stochastic gradient descent (SGD) with oscillating learning rate between $5 \cdot 10^{-5}$ and $15 \cdot 10^{-5}$ as optimizer. This choice is motivated by preliminary studies carried out in \cite{Smith2015}, and experimentally confirmed in our analysis.

\section{Discussion}\label{sec:results}
In this section we discuss the experimental results. First we show the capability of the proposed approach to distinguish between patches that contain camera model information and patches that are not suitable for this task. Then, we show how it is possible to improve camera model identification using the proposed approach.

\subsection{Patch Reliability}
In order to validate the patch reliability estimation we perform a set of tests.

\textbf{CNN Architecture.}
The first set of experiments has been devoted to the choice of a network architecture for $\Md$. To this purpose, we trained a set of different architectures for 15 epochs using the \textit{Pre-Trained} strategy. As architectures we selected all possible combinations of up to six inner product layers (with ReLU activation) composed by 32, 64 or 128 neurons each. The last layer is always set to two neurons followed by softmax. From this experiment, we selected the model $\Md$ with the highest validation accuracy for each tested amount of layers, which are:
\begin{itemize}
	\item $\Md^2$ composed by two inner product layers with 128 and 2 neurons, respectively.
	\item $\Md^3$ composed by three inner product layers with 64, 128 and 2 neurons, respectively.
	\item $\Md^4$ composed by four inner product layers with 64, 32, 128 and 2 neurons, respectively.
	\item $\Md^5$ composed by five inner product layers with 64, 32, 64, 128 and 2 neurons, respectively.
	\item $\Md^6$ composed by six inner product layers with 64, 32, 32, 64, 64 and 2 neurons, respectively.
\end{itemize}

\begin{figure}[t]
	\centering
	\subfloat[(a) Loss]{\includegraphics[width=\columnwidth]{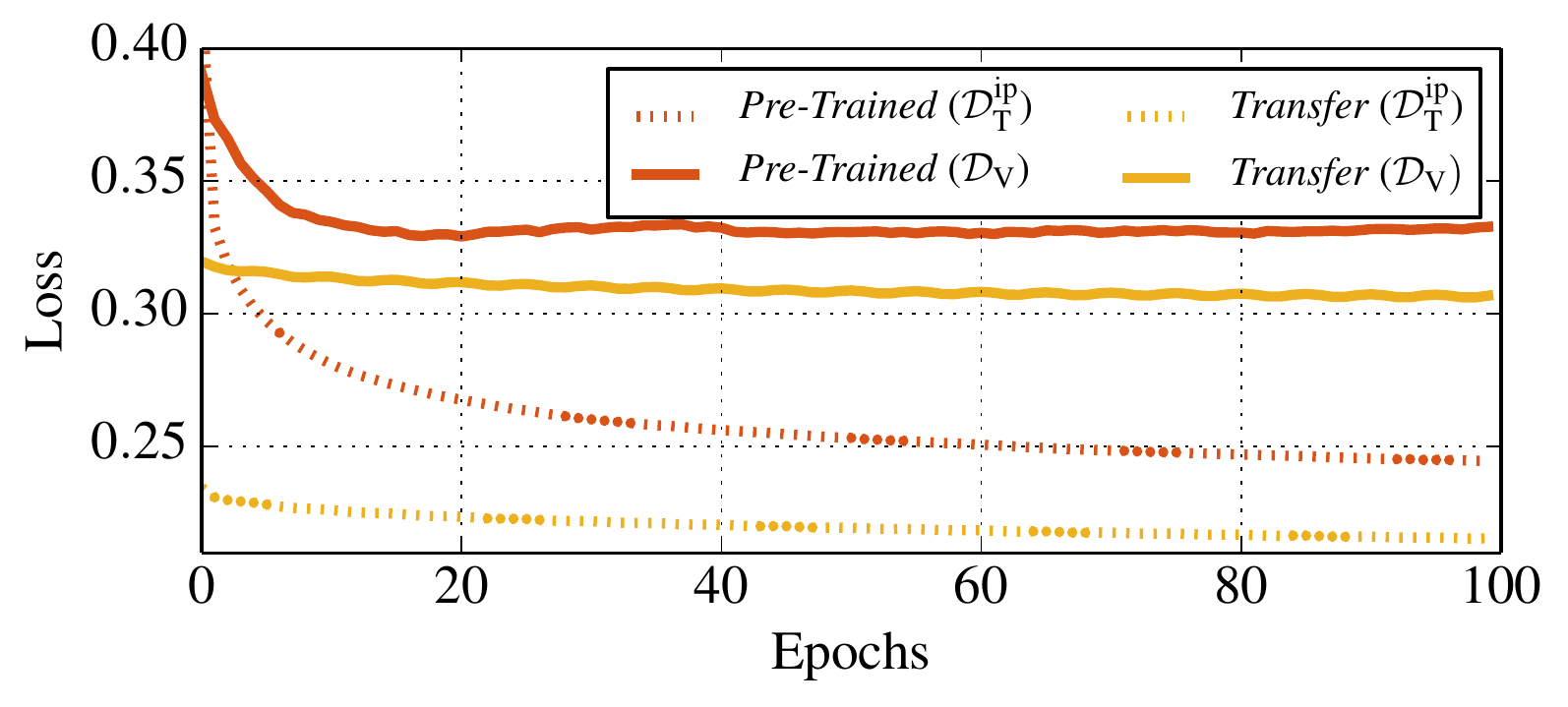}} \\
	\subfloat[(b) Accuracy]{\includegraphics[width=\columnwidth]{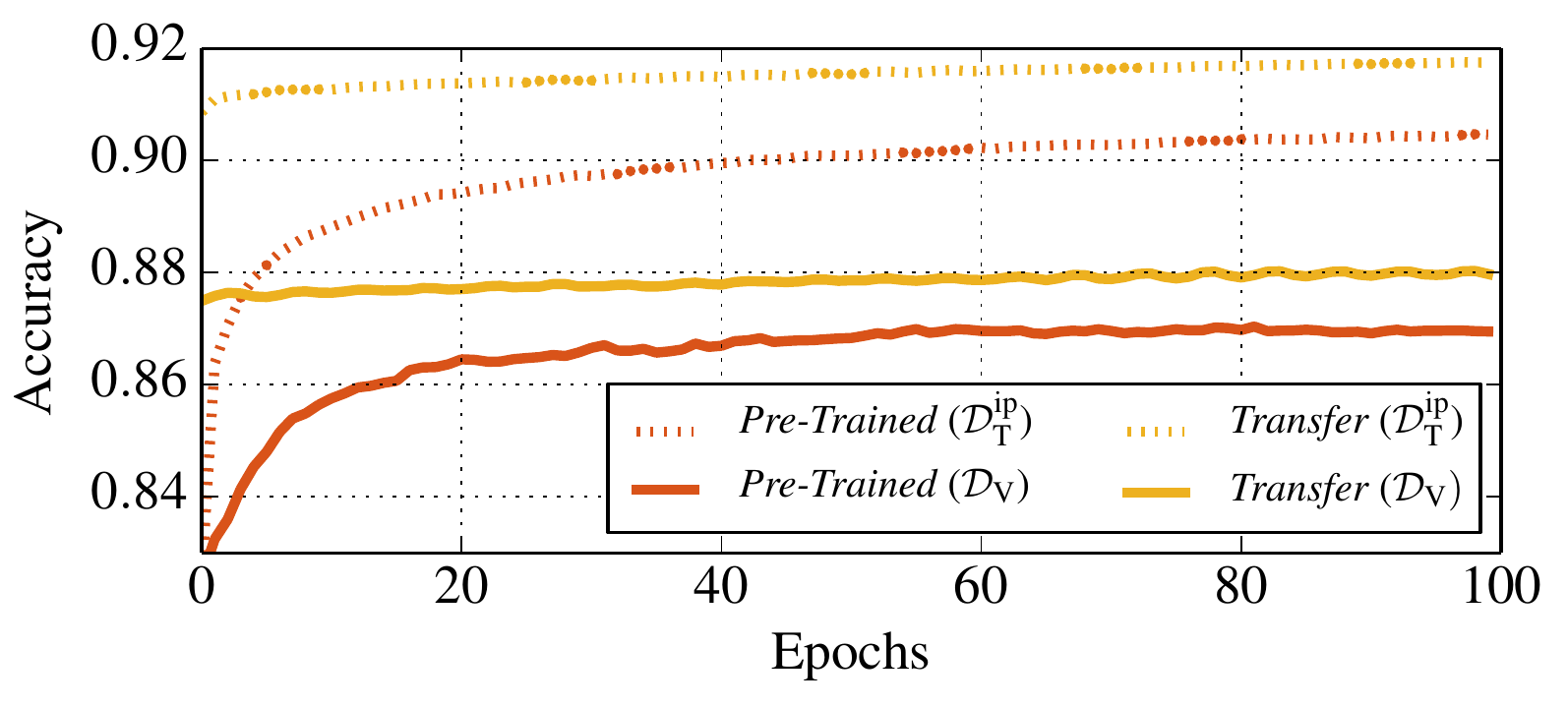}}
	\caption{Loss and accuracy curves on training ($\Dtip$) and validation ($\Dv$) datasets using \textit{Pre-Trained} and \textit{Transfer} strategies on $\Md^4$.}
	\label{fig:curves}
\end{figure}

\textbf{Training Strategy.}
In order to validate the proposed two-tiered \textit{Transfer} training strategy, we trained the five selected models with the \textit{Scratch}, \textit{Pre-Trained} and \textit{Transfer} strategies. Examples of training (on $\Dtip$) and validation (on $\Dv$) loss curves for the \textit{Pre-Trained} and \textit{Transfer} strategies on $\Md^4$ are shown in Figure~\ref{fig:curves}(a). It is possible to notice that the chosen optimizers enable a smooth loss decrease over several epochs. Moreover, the \textit{Transfer} strategy provides a lower loss on both training and validation data, thus yielding better results compared to \textit{Pre-Trained}. Similar conclusions can be drawn from the accuracy curves presented in Figure~\ref{fig:curves}(b). We do not display curves for the \textit{Scratch} strategy, as it is always worse than both the \textit{Pre-Trained} and \textit{Transfer} strategies. This was expected, as the amount of used training data in $\Dtip$ is probably not enough to learn all parameters of $\Mf$. Therefore, starting from a pre-trained $\Mc$ becomes necessary.
\begin{figure}[t]
	\centering
	\includegraphics[width=\columnwidth]{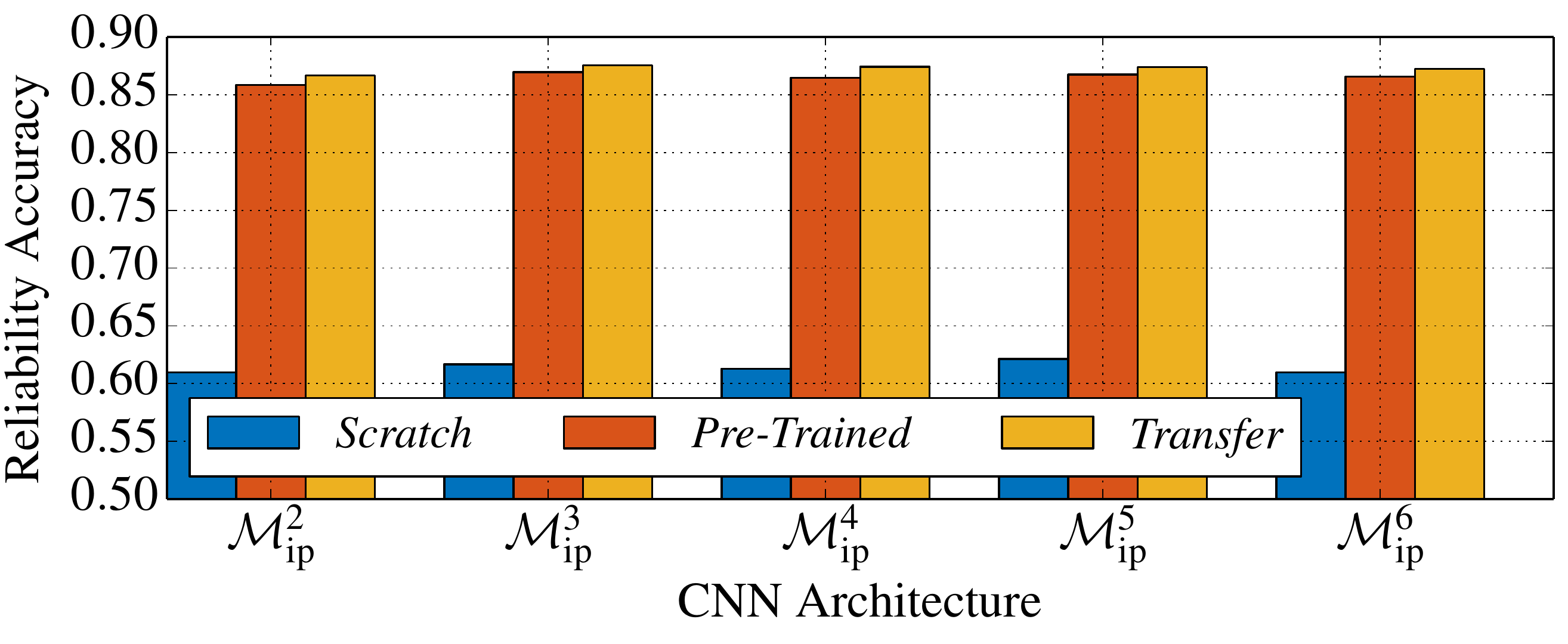}
	\caption{Patch reliability estimation accuracy. The \textit{Transfer} strategy yields the most accurate results for every architecture.}
	\label{fig:rel_accuracy}
\end{figure}
\begin{figure}[t]
	\centering
	\includegraphics[width=\columnwidth]{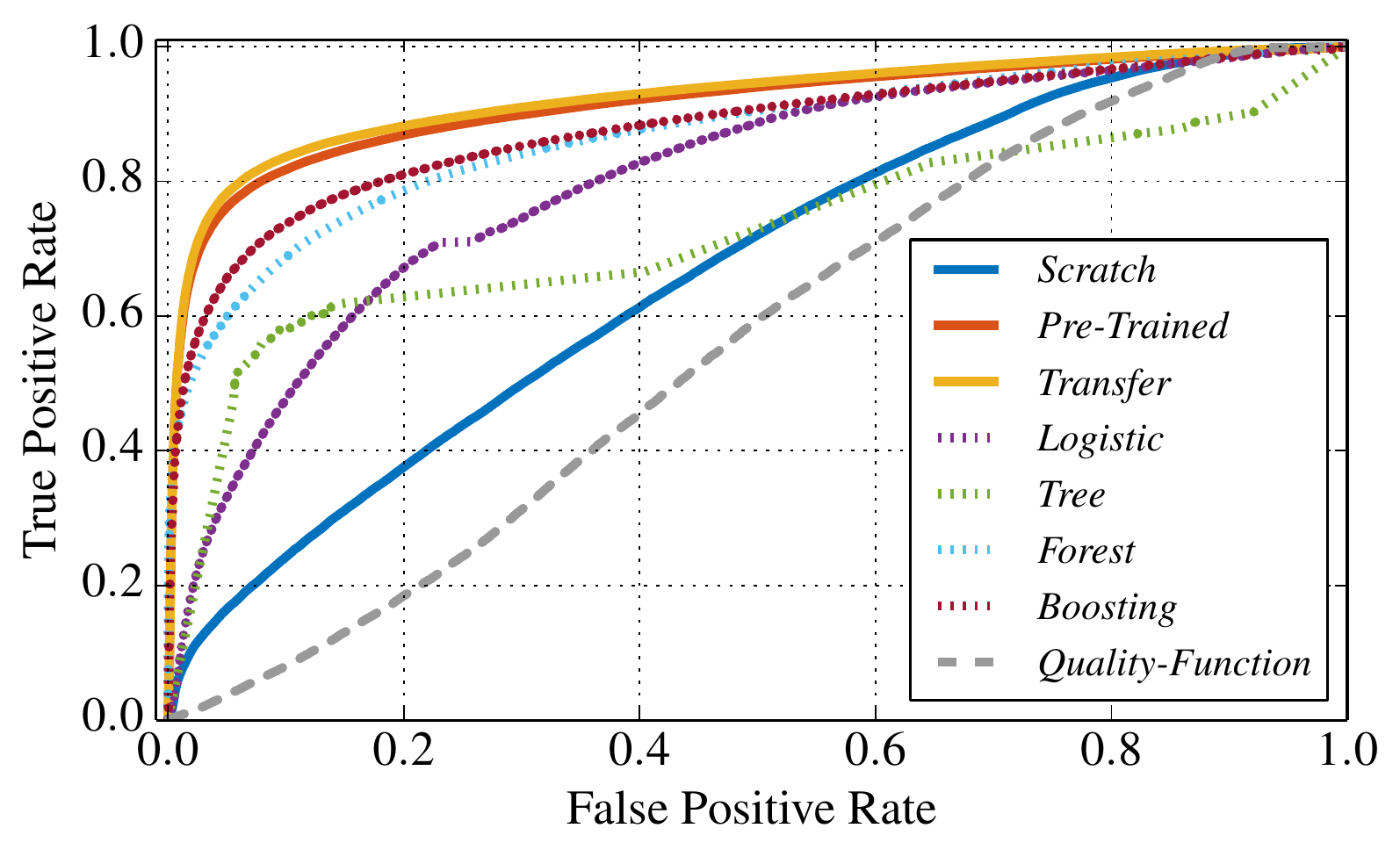}
	\caption{ROC curves on reliable patch detection. The proposed methods are represented by means of solid lines. Other baselines are dotted or dashed. Our \textit{Transfer} strategy achieves the best overall performance.}
	\label{fig:roc}
\end{figure}

Figure~\ref{fig:rel_accuracy} shows the reliability patch estimation accuracy for all models from $\Md^2$ to $\Md^6$, trained with all three training strategies and tested on the evaluation dataset $\De$. For each architecture, we selected the model with highest validation accuracy achieved over 100 epochs. These results further confirm that the \textit{Scratch} strategy is not a viable solution for this problem. The \textit{Transfer} strategy is the best choice for each network, achieving around $86\%$ accuracy in detecting reliable patches. In other words, $86\%$ of the selected patches will be correctly attributed to their camera, whereas only $14\%$ of them will be wrongly classified. In the ideal scenario of errors uniformly spread across all models and images, this means we could use a majority voting strategy to further increase accuracy at the image level. 

From Figure~\ref{fig:rel_accuracy}, it is also possible to notice that increasing $\Md$ depth does not increase accuracy. Therefore, from this point on, we only consider architecture $\Md^4$ as a good trade-off.

\textbf{Baselines Comparison.}
In order to further validate the proposed approach, we also considered two possible baseline solutions.

The first one consists in using other kinds of supervised classifiers exploiting the 18-element vector returned by $\Mc$ as feature. To this purpose, we trained a logistic regressor (\textit{Logistic}), a decision tree (\textit{Tree}), a random forest (\textit{Forest}) and a gradient boosting classifier (\textit{Boosting}). For each method, we applied z-score feature normalization and we selected the model achieving highest validation accuracy on $\Dv$ after a parameter grid-search training on $\Dtip$. Accuracy results on patch reliability on evaluation set $\De$ were $70.7\%$, $73.9\%$, $78.6\%$ and $81.8\%$, respectively. None of them approaches the $86\%$ of the proposed solution.

The second baseline solution we tested is the quality-function presented in \cite{Bondi2017a} (\textit{Quality-Function}). This function is computed for each patch and returns a value between zero and one indicating whether the patch is suitable for training $\Mc$. Although \text{Quality-Function} was not intended to work as test reliability indicator, we decided that a comparison was necessary for completeness. To this purpose, Figure~\ref{fig:roc} shows receiver operating characteristic (ROC) curves obtained thresholding our reliability likelihood estimation $g_k$, the soft output of the other classifiers (i.e., logistic regressor, decision tree, etc.), and the quality-function returned value \cite{Bondi2017a}. As expected, the use of the quality-function presented in \cite{Bondi2017a} provides less accurate results. Conversely, the proposed method achieves better performance than all other classifiers when trained according to \textit{Transfer} strategy.

\begin{table}[]
	\centering
	\caption{Camera model attribution accuracy using selected reliable patches from test dataset only. Using \textit{Transfer} strategy (bold), the amount of selected patches in $\De$ is always greater than $77\%$ of $|\De|$. Accuracy improvement over random patch selection is greater than $8\%$.}
	\label{tab:camera_model_accuracy}
	\resizebox{\columnwidth}{!}{
		\begin{tabular}{cc|c|c|c}
			\hline
			\textbf{$\Md$} & \textbf{Strategy}    & \textbf{Patches} & \textbf{Accuracy} & \textbf{Acc. Delta}   \\ \hline
			& \textit{Scratch}     & 553 475           & 0.9009            & 0.0342          \\
			$\Md^2$        & \textit{Pre-Trained} & 618 958           & 0.9478            & 0.0811          \\
			& \textit{Transfer}    & 637 135  & \textbf{0.9513}   & \textbf{0.0845} \\ \hline
			& \textit{Scratch}     & 518 228           & 0.9041            & 0.0374          \\
			$\Md^3$        & \textit{Pre-Trained} & 626 767          & 0.9520            & 0.0853          \\
			& \textit{Transfer}    & 641 808  & \textbf{0.9556}   & \textbf{0.0888} \\ \hline
			& \textit{Scratch}     & 562 897           & 0.8963            & 0.0296          \\
			$\Md^4$        & \textit{Pre-Trained} & {649 515}  & 0.9499            & 0.0832          \\
			& \textit{Transfer}    & 647 998          & \textbf{0.9532}   & \textbf{0.0865} \\ \hline
			& \textit{Scratch}     & 511 425           & 0.9045            & 0.0378          \\
			$\Md^5$        & \textit{Pre-Trained} & 648 665           & 0.9529            & 0.0862          \\
			& \textit{Transfer}    & 651 508  & \textbf{0.9530}   & \textbf{0.0863} \\ \hline
			& \textit{Scratch}     & 517 386           & 0.9035            & 0.0367          \\
			$\Md^6$        & \textit{Pre-Trained} & 651 405           & 0.9501            & 0.0834          \\
			& \textit{Transfer}    & 652 308  & \textbf{0.9531}   & \textbf{0.0864} \\ \hline
		\end{tabular}
	}
\end{table}

\subsection{Camera Model Attribution}
After validating the possibility of selecting reliable patches with the proposed method, we tested the effect of this solution on camera model attribution. To this purpose, we report in Table~\ref{tab:camera_model_accuracy} the evaluation set results for the five investigated $\Md$ models and the three training strategies when a single patch is used for camera model attribution. We do so in terms of:
\begin{enumerate}
	\item \textit{Patches}, i.e., the number of estimated reliable patches.
	\item \textit{Accuracy}, i.e., the average achieved camera model attribution result.
	\item \textit{Accuracy Delta}, i.e., the accuracy increment with respect to not using patch selection but randomly picking them (i.e., \cite{Bondi2017}).
\end{enumerate}
These results highlight that it is possible to improve camera model attribution by more than $8\%$.

\begin{figure}[t]
	\centering
	\includegraphics[width=\columnwidth]{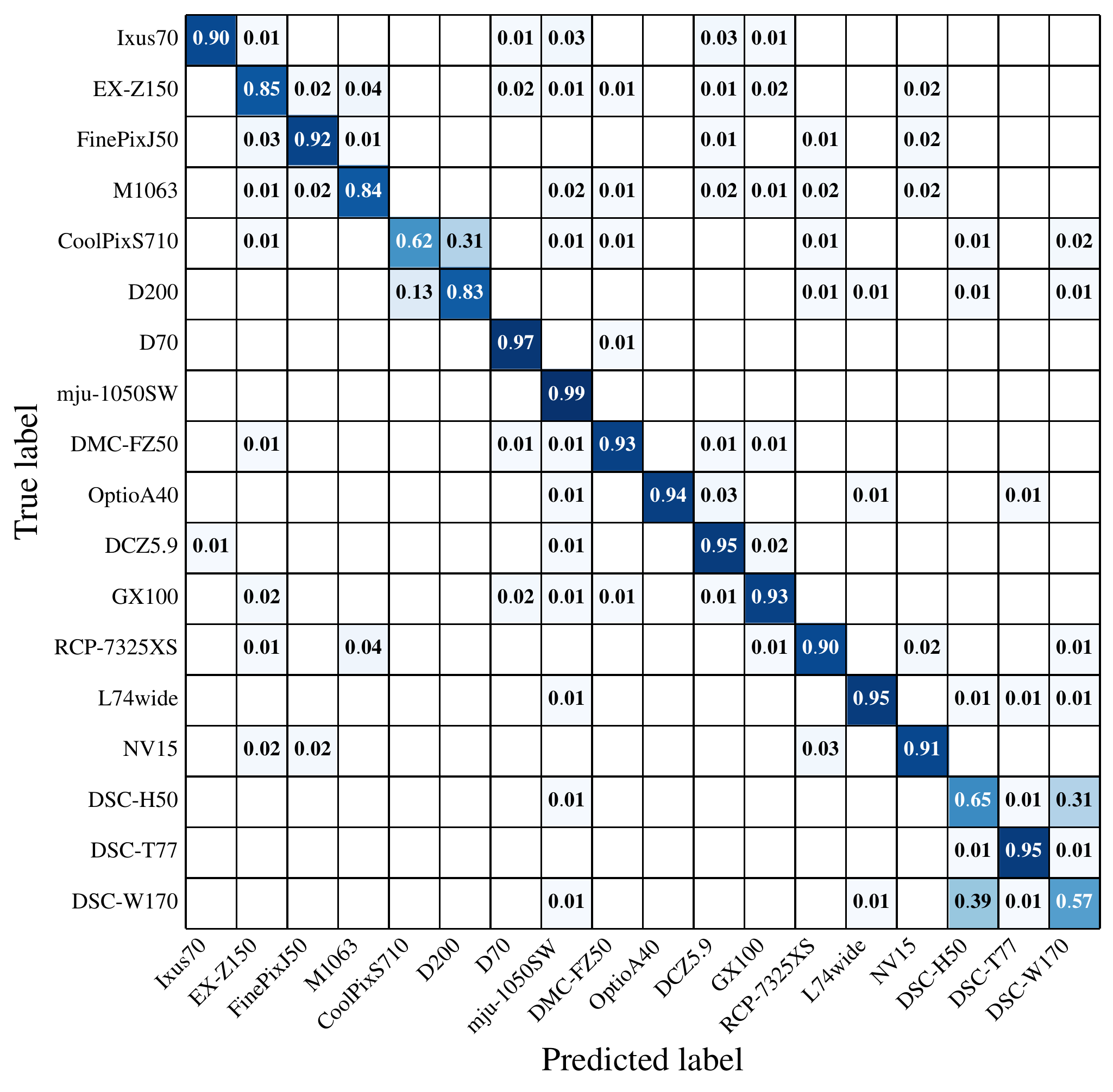}
	\caption{Camera model attribution confusion matrix obtained with $\Mc$ on $\De$ without patch selection.}
	\label{fig:cm_before}
\end{figure}

\begin{figure}[t]
	\centering
	\includegraphics[width=\columnwidth]{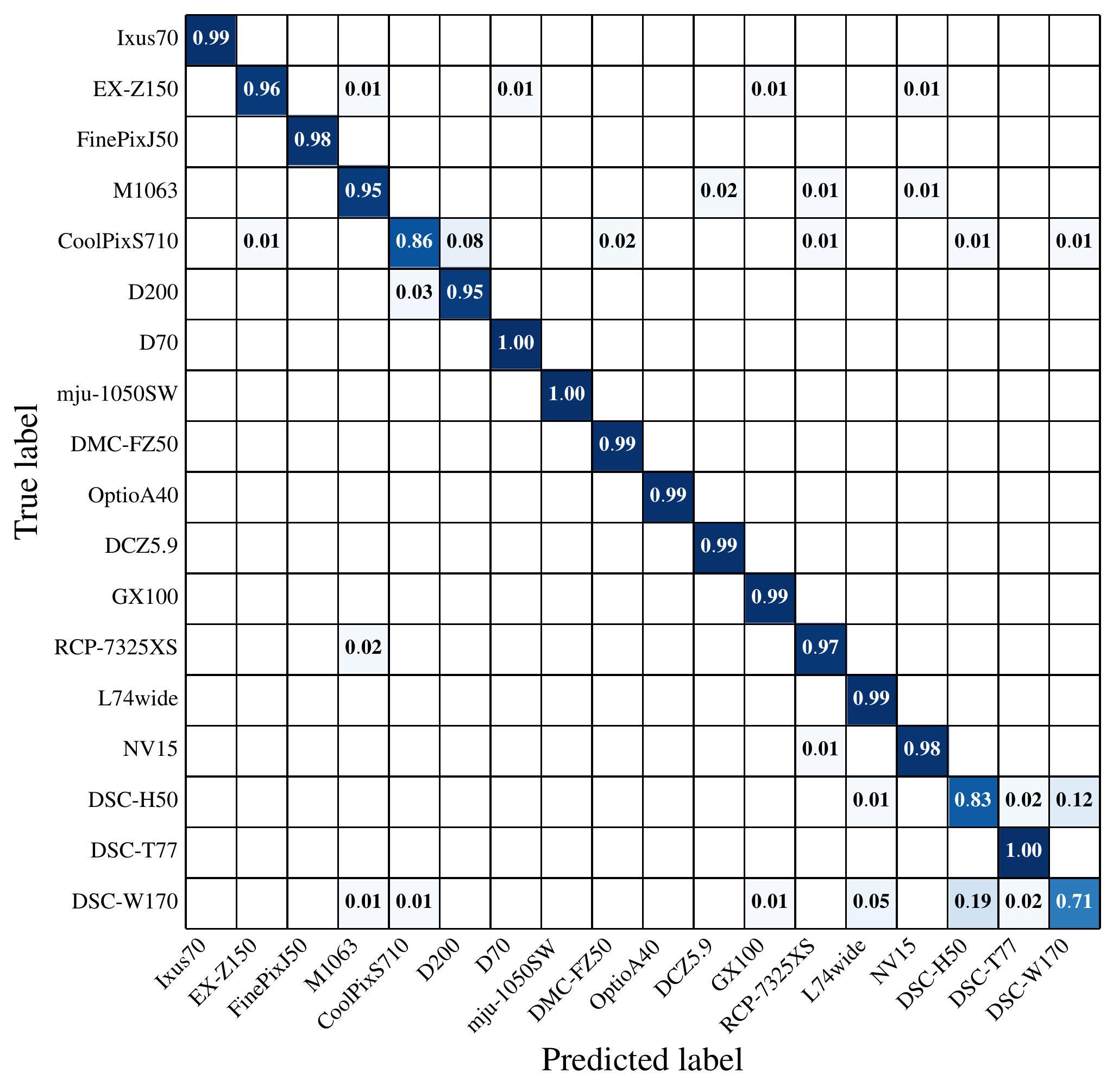}
	\caption{Camera model attribution confusion matrix obtained with $\Mc$ on $\De$ using patches selected with $\Md^4$.}
	\label{fig:cm_after}
\end{figure}

Figure~\ref{fig:cm_before} shows confusion matrix results using $\Mc$ (i.e., the output of the network proposed in \cite{Bondi2017}) on evaluation data $\De$ randomly selecting patches. The average accuracy per patch is $87\%$. Figure~\ref{fig:cm_after} shows the same results, evaluating only patches considered reliable using $\Md^4$. In this scenario, accuracy increases to more than $95\%$. By comparing the two figures, it is possible to notice that many spurious classifications outside of the confusion matrix diagonal are corrected by the use of reliable patches only.


\section{Conclusions}\label{sec:conclusions}
In this paper we presented a method for reliability patch estimation for camera model attribution. This means being able to estimate the likelihood that an image patch will be correctly attributed to the camera model used to acquire the image from which the patch comes from.

The proposed solution is based on concatenating a pre-trained CNN for patch-wise camera model attribution with a dense network that acts as a binary classifier. Exploiting transfer learning techniques and a two-tiered training strategy, it is possible to achieve $86\%$ of accuracy in patch reliability estimation. Moreover, by running camera model attribution on single selected patches, camera attribution accuracy increases by more than $8\%$.

In addition to the impact on camera model attribution, the proposed method returns a reliability mask that highlights which image regions are considered reliable in terms of camera attribution. This could be useful in the future to better understand which image details are more important to camera model attribution CNNs. Moreover, it could be paired with splicing localization algorithms based on camera model traces to possibly opt-out unreliable regions from the analysis. 

\section*{Acknowledgments}\label{sec:acknowledgment}
This material is based on research sponsored by the Defense Advanced Research Projects Agency (DARPA) and the Air Force Research Laboratory (AFRL) under agreement number FA8750-16-2-0173.  The U.S. Government is authorized to reproduce and distribute reprints for Governmental purposes notwithstanding any copyright notation thereon. The views and conclusions contained herein are those of the authors and should not be interpreted as necessarily representing the official policies or endorsements, either expressed or implied, of DARPA, AFRL or the U.S. Government.

{\small
\bibliographystyle{ieee}
\bibliography{biblio}
}

\end{document}